\pdfoutput=1

\documentclass[11pt]{article}

\usepackage{EMNLP2022}

\usepackage{times}
\usepackage{latexsym}

\usepackage[T1]{fontenc}

\usepackage[utf8]{inputenc}

\usepackage{microtype}

\usepackage{inconsolata}

\usepackage{booktabs}
\usepackage{multirow}
\usepackage{graphicx}
\usepackage{subcaption}

\title{Does Corpus Quality Really Matter for Low-Resource Languages?}

\author{\ Mikel Artetxe$^{1}$ \qquad Itziar Aldabe$^{2}$ \qquad Rodrigo Agerri$^{2}$ \\
{\bf Olatz Perez-de-Viñaspre$^{2}$} \qquad {\bf Aitor Soroa$^{2}$} \\
$^1$Meta AI \\ $^2$HiTZ Center, University of the Basque Country (UPV/EHU) \\
\texttt{artetxe@meta.com} \\
\texttt{\{itziar.aldabe,rodrigo.agerri,olatz.perezdevinaspre,a.soroa\}@ehu.eus}
}

\begin{document}
\maketitle
\begin{abstract}
The vast majority of non-English corpora are derived from automatically filtered versions of CommonCrawl. While prior work has identified major issues on the quality of these datasets \citep{kreutzer2021quality}, it is not clear how this impacts downstream performance. Taking representation learning in Basque as a case study, we explore tailored crawling---manually identifying and scraping websites with high-quality content---as an alternative to filtering CommonCrawl. Our new corpus, called EusCrawl, is similar in size to the Basque portion of popular multilingual corpora like CC100 and mC4, yet it has a much higher quality according to native annotators. For instance, $66\%$ of documents are rated as high-quality for EusCrawl, in contrast with $<33\%$ for both mC4 and CC100. Nevertheless, we obtain similar results on downstream NLU tasks regardless of the corpus used for pre-training. Our work suggests that NLU performance in low-resource languages is not primarily constrained by the quality of the data, and other factors like corpus size and domain coverage can play a more important role.
\end{abstract}

\section{Introduction}

Large-scale pre-training has resulted in a paradigm shift in NLP \citep{bommasani2021opportunities}. While recent progress has been primarily driven by scaling up on model size and compute, both data quantity and quality have been shown to play a critical role \citep{kaplan2020scaling,rae2022scaling}. %
Nevertheless, existing efforts on data curation have primarily focused on English, and recent work on multilingual pre-training has relied on automatically filtered versions of CommonCrawl. For instance, XLM-R was trained on CC100 \citep{conneau-etal-2020-unsupervised}, mT5 was trained on mC4 \citep{xue-etal-2021-mt5}, and XGLM was trained on CC100-XL \citep{lin2021fewshot}, which were all obtained by running language identification on several CommonCrawl snapshots and filtering through language-agnostic approaches. Unfortunately, \citet{kreutzer2021quality} identified major issues on the quality of such multilingual datasets, ranging from language identification errors to boilerplate and non-linguistic content. However, the practical impact of these issues has not been studied, and it is unclear the extent to which higher-quality data could lead to better performance in low-resource languages.

In this paper, we take representation learning in Basque as a case study, and explore tailored crawling (i.e., manually identifying and scraping websites with high-quality content) as an alternative to filtering CommonCrawl. We introduce EusCrawl, a new corpus for Basque comprising 12.5M documents from 33 websites with Creative Commons content. EusCrawl is similar in size to the Basque portion of CC100 and mC4, but it has substantially less issues and a higher perceived quality according to our blind audit with native annotators. However, we find that this improvement does not carry over to downstream NLU tasks, as masked language models pre-trained on either corpora obtain similar results on 5 benchmarks. Our results suggests that data quantity and domain coverage play a more important role, prompting for methods to exploit more diverse sources of data in low-resource languages.

\begin{table*}[ht]
\begin{center}
\begin{small}
\begin{tabular}{rrrrl}
\toprule
& Size & Tokens & Docs & Source \\
\midrule
mC4 \citep{xue-etal-2021-mt5} & 4,387 MiB & 1,004M & 30,098k & Filtered CommonCrawl \\
CC100 \citep{conneau-etal-2020-unsupervised} & 2,027 MiB & 416M & 16,761k & Filtered CommonCrawl \\
Wikipedia & 313 MiB & 66M & 2,685k & Wikipedia dump \\
\midrule
EusCrawl (ours) & 2,149 MiB & 423M & 12,528k & Tailored crawling (see Table \ref{tab:euscrawl_subsets})\\
\bottomrule
\end{tabular}%
\end{small}
\end{center}
\caption{Basque corpora used in our experiments. We report uncompressed text size, number of SentencePiece tokens (using a 50K vocabulary learned in each corpus), and number of documents.}
\label{tab:corpora}
\end{table*}

\begin{table*}[ht]
\begin{center}
\begin{small}
\begin{tabular}{rrrrll}
\toprule
& Size & Tokens & Docs & License & Domain \\
\midrule

\href{https://tokikom.eus/}{Tokikom}$^\dagger$ & 784 MiB & 153M & 4,961k & CC-BY-SA & Local media \\

\href{https://www.berria.eus/}{Berria} & 525 MiB & 101M & 2,193k & CC-BY-SA & National newspaper \\

\href{https://hitza.eus/}{Hitza}$^\ddagger$ & 418 MiB & 80M & 2,257k & CC-BY-NC-ND & Regional newspapers \\

\href{https://eu.wikipedia.org}{Wikipedia} & 313 MiB & 68M & 2,685k & CC-BY-SA & Encyclopedia \\

\href{https://www.argia.eus/}{Argia} & 101 MiB & 20M & 370k & CC-BY-SA & News magazine \\

\href{https://www.bilbohiria.eus/}{Bilbo Hiria irratia} & 7 MiB & 1M & 54k & CC-BY-NC-SA & Radio station\\
\href{https://www.sarean.eus/}{Sarean} & 2 MiB & 0.3M & 8k & CC-BY-SA & Technology blog \\

\bottomrule
\end{tabular}%
\end{small}
\end{center}
\caption{
Data sources used to build EusCrawl.
$^\dagger$Tokikom is a network of local media; we include \href{https://aiaraldea.eus/}{Aiaraldea}, \href{https://aikor.eus/}{Aikor}, \href{https://anboto.org/}{Anboto}, \href{https://ataria.eus/}{Tolosaldeko Ataria}, \href{https://aiurri.eus/}{Aiurri}, \href{https://erran.eus/}{Erran}, \href{https://euskalerriairratia.eus/}{Euskalerria Irratia}, \href{https://goiena.eus/}{Goiena}, \href{https://guaixe.eus/}{Guaixe}, \href{https://hiruka.eus/}{Hiruka}, \href{https://karkara.eus/}{Karkara}, \href{https://maxixatzen.eus/}{Maxixatzen}, \href{https://plaentxia.eus/}{Plaentxia}, \href{https://alea.eus/}{Alea}, \href{https://noaua.eus/}{Noaua}, \href{https://txintxarri.eus/}{Txintxarri}, \href{https://uztarria.eus/}{Uztarria}, \href{https://amezti.eus/}{Amezti}, \href{https://zarauzkohitza.eus/}{Zarauzko Hitza}, \href{https://kronika.eus/}{Kronika} and \href{https://www.geuria.eus/}{Geuria}.
$^\ddagger$Hitza is a family of regional newspapers; we include \href{https://bidasoa.hitza.eus}{Bidasoko Hitza}, \href{https://busturialdea.hitza.eus}{Busturialdeko Hitza}, \href{https://goierri.hitza.eus}{Goierriko Hitza}, \href{https://irutxulo.hitza.eus}{Irutxuloko Hitza}, \href{https://lea-artibaietamutriku.hitza.eus}{Lea-Artibai eta Mutrikuko Hitza}, \href{https://oarsoaldea.hitza.eus}{Oarsoaldeko Hitza} and \href{https://urolakosta.hitza.eus/}{Urola Kostako Hitza}.
}
\label{tab:euscrawl_subsets}
\end{table*}

This paper makes the following contributions:
(i) we release EusCrawl, a high-quality corpus for Basque comprising 12.5M documents and 423M tokens;\footnote{{\scriptsize{\url{https://www.ixa.eus/euscrawl/}}}. Meta AI was not involved in the collection and distribution of the corpus.}
(ii) we manually assess the quality of EusCrawl in comparison with mC4 and CC100, finding that it has substantially less issues and a higher perceived quality according to native annotators;
(iii) we compare masked language models pre-trained on EusCrawl, mC4, CC100 and Wikipedia\footnote{Models available at {\scriptsize{\texttt{https://dl.fbaipublicfiles.com/ euscrawl/roberta-eus-\{euscrawl|mc4|cc100|wikipedia\}- \{base|large\}.tar.gz}}}.} on 5 NLU tasks, finding that they all perform similarly with the exception of Wikipedia; and
(iv) we obtain state-of-the-art results on several NLU benchmarks in Basque, outperforming prior work that relied on non-public corpora.

\section{Experimental setup}

We next detail the corpora compared in our experiments (\S\ref{subsec:corpora}), and the qualitative and downstream evaluation settings (\S\ref{subsec:qualitative_settings} and \S\ref{subsec:downstream_settings}).

\subsection{Corpora} \label{subsec:corpora}

We compare 4 Basque corpora in our experiments: mC4, CC100, Wikipedia and EusCrawl. Table \ref{tab:corpora} summarizes their details.
\textbf{mC4}\footnote{We use the version released by AllenAI at \scriptsize{\url{https://github.com/allenai/allennlp/discussions/5265}}} and \textbf{CC100}\footnote{We use the version from \scriptsize{\url{https://data.statmt.org/cc-100/}}} are, to the best of our knowledge, the two largest public corpora for Basque. They were introduced to train mT5 \citep{xue-etal-2021-mt5} and XLM-R \citep{conneau-etal-2020-unsupervised}, respectively, and were built by filtering CommonCrawl.
\textbf{Wikipedia} has been a popular source for multilingual data \citep{pires-etal-2019-multilingual,conneau2019xlm,artetxe-etal-2020-cross}. We extract text from a Wikipedia dump using the WikiExtractor tool.\footnote{\scriptsize{\url{https://github.com/attardi/wikiextractor}}}
\textbf{EusCrawl} is a new corpus we introduce. Instead of filtering CommonCrawl, we do tailored crawling on 33 websites with high-quality content in Basque, mostly on the news domain. We build ad-hoc scrapers to extract text from these websites, resulting in higher coverage\footnote{While one may expect the websites we crawl to be covered by mC4 and CC100, a large fraction of this content is missing in them. This is both because CommonCrawl is far from being a complete dump of the Internet, and the filtering applied by CC100 and mC4 is noisy, removing valid content.} and cleaner text compared to general purpose approaches. We only use content with a Creative Commons license. Table \ref{tab:euscrawl_subsets} summarizes all the sources we use.

\begin{figure*}[t]
     \centering
     \begin{subfigure}[b]{0.485\textwidth}
         \centering
         \includegraphics[width=\textwidth]{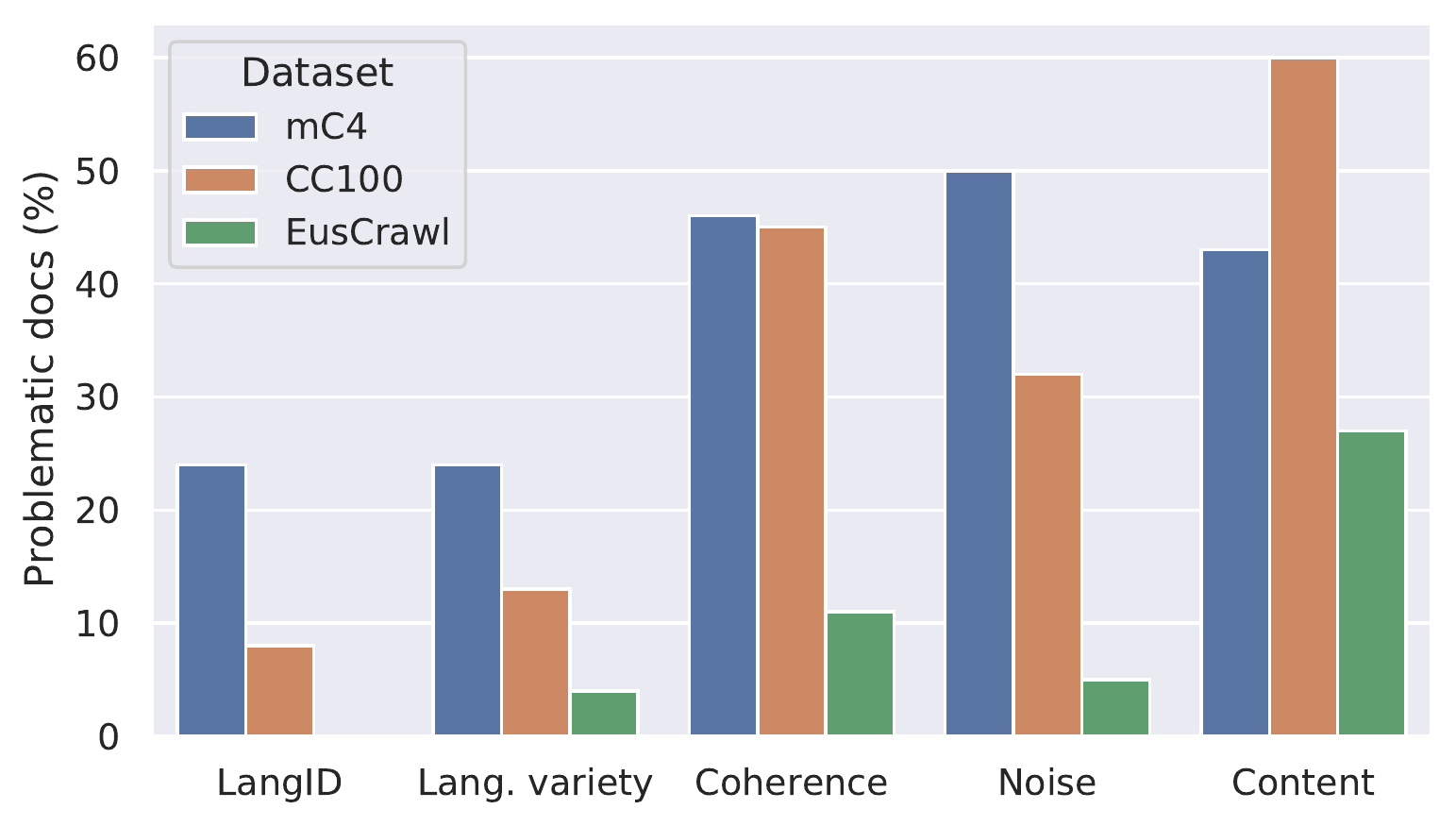}
         \caption{Issues}
         \label{fig:issues}
     \end{subfigure}
     \hfill
     \begin{subfigure}[b]{0.485\textwidth}
         \centering
         \includegraphics[width=\textwidth]{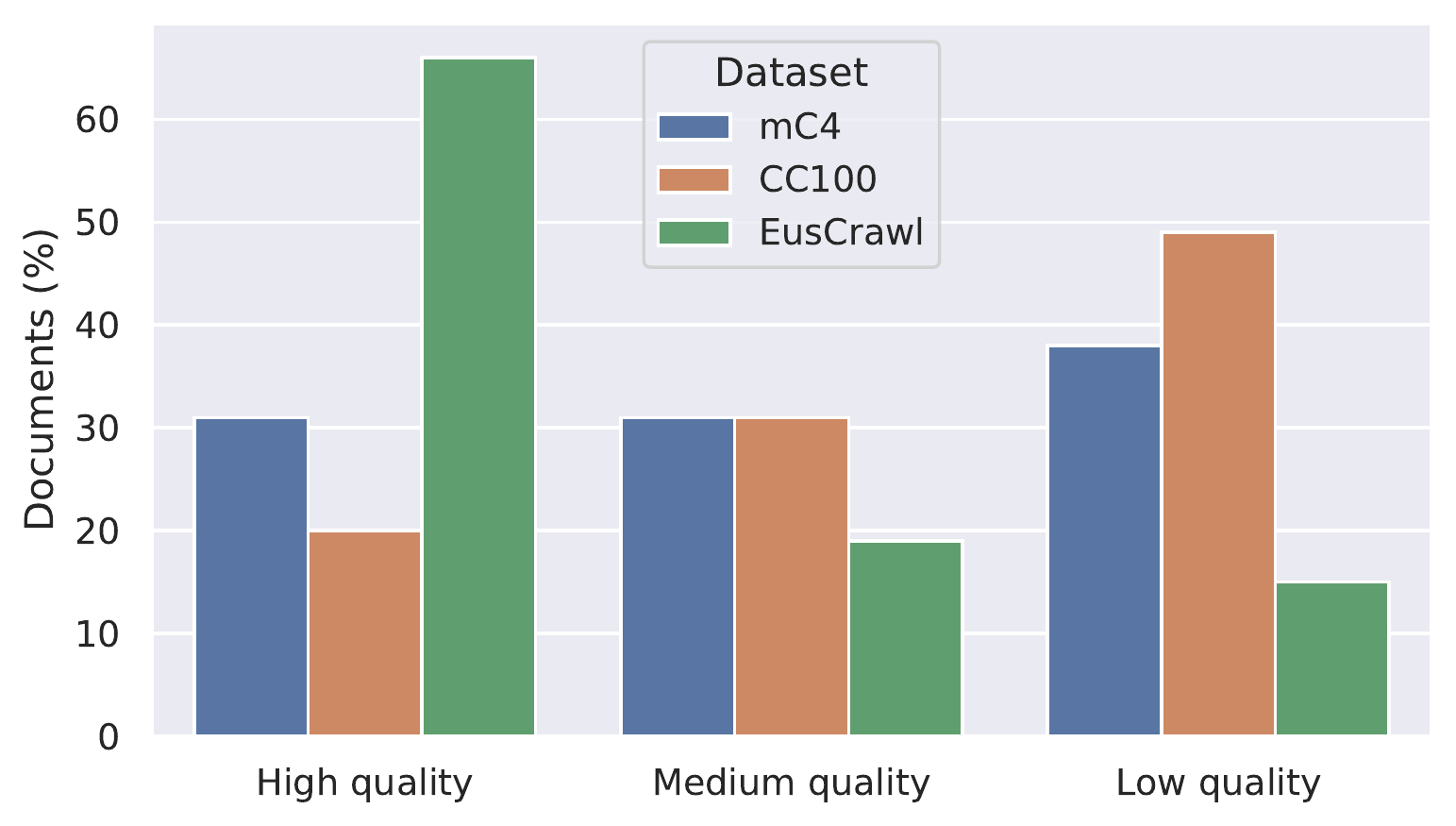}
         \caption{Overall quality}
         \label{fig:quality}
     \end{subfigure}
     \hfill
    \caption{Data audit results. EusCrawl has a much higher quality than mC4 and CC100. See \S\ref{subsec:qualitative_settings} for more details.}
    \label{fig:qualitative}
\end{figure*}

\subsection{Qualitative evaluation} \label{subsec:qualitative_settings}

We manually audit the quality of EusCrawl in comparison with mC4 and CC100 by randomly sampling 100 documents from each corpus (a total of 300 documents), and asking native annotators to assess their quality.\footnote{
So as to control for the variance across annotators, we asked two additional native speakers to evaluate a random subset of 100 documents. The main findings were consistent across all the 3 annotations, so we omit results for brevity.
} We ensure that the evaluation is blind by showing the documents in a random order and not revealing what corpus they were sampled from. For each document, we ask the annotators to assess if the document has any problem in each of the following categories: \textbf{langID} (the document is not in Basque), \textbf{language variety} (the document is not written in standard and correct Basque), \textbf{coherence} (the document has gaps and/or some portions are not connected), \textbf{noise} (the document is not clean) and \textbf{content} (the document seems to have been generated automatically and/or has no meat). In addition, we ask annotators to classify each document according to its \textbf{perceived quality} as high-quality (the document does not have quality issues and the annotator thinks that it would be good to have it in the corpus), medium-quality (the document has some minor issues and the annotator is unsure if it would be good to have it in the corpus), or low-quality (the document has major issues and the annotator thinks that it would be better not to have it in the corpus). Refer to Appendix \ref{app:instructions} for the complete instructions given to annotators.

\subsection{Downstream evaluation} \label{subsec:downstream_settings}

In addition to the qualitative evaluation, we pre-train RoBERTa models \citep{liu2019roberta} on each corpus, and evaluate fine-tuning them on the following \textbf{NLU benchmarks}: topic classification on BHTC \citep{agerri-EtAl:2020:LREC}, sentiment classification on Behagune \citep{agerri-EtAl:2020:LREC}, stance detection on VaxxStance \cite{vaxxstance2021}, Named Entity Recognition (NER) on EIEC \citep{alegria2006lessons}, and extractive conversational Question Answering (QA) on Elkarrizketak \cite{otegi-EtAl:2020:LREC}. We provide additional details on these datasets in Appendix \ref{app:downstream_settings}.

We \textbf{pre-train} each model for 125k steps with a batch size of 2048 and a sequence length of 512, using the same hyperparameters as \citet{liu2019roberta}. We train RoBERTa-base models for our main comparison using a learning rate of 7e-4, and further train a RoBERTa-large model on EusCrawl with a learning rate of 4e-4 to understand the effect of scaling. In all cases, we use the final checkpoint without early stopping. We use SentencePiece \citep{kudo-richardson-2018-sentencepiece} for tokenization, using a 50k vocabulary learned in each separate corpus.

For \textbf{fine-tuning}, we use the same hyperparameters as \citet{agerri-EtAl:2020:LREC}. For topic classification, sentiment classification and stance detection, we use a batch size of 16, a learning rate of 2e-5 with linear decay and a warmup of 6\%, and train the model for 10 epochs. For NER and QA, we use a batch size of 32, a constant learning rate of 5e-5, and train for 4 epochs. We did not perform any hyperparameter tuning or model selection, and report results on the test set. The development sets, when available, were not used.

\begin{table*}[ht]
\begin{center}
\begin{small}
\addtolength{\tabcolsep}{-1.5pt}
\begin{tabular}{lrlllll|l}
\toprule
                  &                               & Topic class. & Sentiment    & Stance det. & NER              & QA         & Avg \\
\midrule
\multirow{3}{*}{\shortstack{Prior best}}
                  & \citet{agerri-EtAl:2020:LREC} & 76.8         & 78.1         & --          & 87.1             & --         & --          \\
                  & \citet{otegi-EtAl:2020:LREC}  & --           & --           & --          & --               & 35.0    & --             \\
                  & \citet{Lai2021WordUpAV}        & --           & --           & 57.3$^{\dagger}$          & --  & --     & --              \\
\midrule
\multirow{4}{*}{\shortstack{RoBERTa-base}}
                  & mC4                           & 75.3\scriptsize{ \textpm 0.7}   & \underline{\textbf{80.4}}\scriptsize{ \textpm 1.5}   & 59.1\scriptsize{ \textpm 5.2}  & 86.0\scriptsize{ \textpm 1.0}       & 35.2\scriptsize{ \textpm 1.8} & 67.2    \\
                  & CC100                         & \underline{76.2}\scriptsize{ \textpm 0.4}   & 78.8\scriptsize{ \textpm 1.2}   & \underline{\textbf{63.4}}\scriptsize{ \textpm 3.5}  & 85.2\scriptsize{ \textpm 1.2}       & \underline{35.8}\scriptsize{ \textpm 1.1} & \underline{67.9}     \\
                  & Wikipedia                     & 70.0\scriptsize{ \textpm 0.8}   & 72.4\scriptsize{ \textpm 2.3}   & 53.2\scriptsize{ \textpm 4.6}  & 71.6\scriptsize{ \textpm 13.1}      & 27.4\scriptsize{ \textpm 0.2} & 58.9     \\
                  & EusCrawl                      & \underline{76.2}\scriptsize{ \textpm 0.6}   & 77.7\scriptsize{ \textpm 1.4}   & 57.4\scriptsize{ \textpm 4.7}  & \underline{86.8}\scriptsize{ \textpm 0.6}       & 34.6\scriptsize{ \textpm 1.8} & 66.5     \\
\midrule
RoBERTa-large     & EusCrawl                      & \textbf{77.6}\scriptsize{ \textpm 0.5}   & 78.8\scriptsize{ \textpm 0.9}   & 62.9\scriptsize{ \textpm 2.3}  & \textbf{87.2}\scriptsize{ \textpm 0.4}       & \textbf{38.3}\scriptsize{ \textpm 1.3} & \textbf{69.0}   \\
\bottomrule
\end{tabular}%
\end{small}
\end{center}
\caption{
Downstream results. We report average F1 and standard deviation across 5 runs (micro F1 in all tasks except stance detection, where we report macro F1 of the \textit{favor} and \textit{against} classes following common practice). $\dagger$Best result among systems that rely exclusively on textual data.
}
\label{tab:downstream}
\end{table*}

\section{Results}

\subsection{Qualitative evaluation}

As shown in Figure \ref{fig:qualitative}, EusCrawl has the best quality by a large margin in all the axes that we consider. mC4 has a slightly higher perceived quality and less content-related issues than CC100, but more problematic documents in the other categories.

More concretely, we find that both mC4 and CC100 have a high proportion of documents with coherence, noise and content-related issues. In addition, mC4 has a significant number of langID and language variety problems. In contrast, EusCrawl has minimal issues in all categories but content, where it still does substantially better than mC4 and CC100. Taking a closer look, we find that most of these content-related issues in EusCrawl correspond to short, template-based Wikipedia articles (e.g., \textit{Placosoma is a a genus of lizards in the family Gymnophthalmidae. They live in Brazil.}\footnote{Original text in Basque: \textit{Placosoma Gymnophthalmidae familiako narrasti genero bat da. Brasilen bizi dira.}}), which should be easy to filter in future iterations. Finally, we find that the overall quality of EusCrawl documents is also much better according to native annotators, with approximately two thirds of the documents being annotated as high-quality, compared to less than one third for both mC4 and CC100.

All in all, our qualitative evaluation provides further evidence that multilingual corpora derived from CommonCrawl have major quality issues, and shows that tailored crawling can be an effective alternative to obtain high-quality data.

\subsection{Downstream tasks}

We report our downstream results in Table \ref{tab:downstream}.

In contrast with the qualitative evaluation, we find that there is not a clear winner among mC4, CC100 and EusCrawl. In fact, when looking at RoBERTa-base results, we find that mC4 does the best on sentiment classification, CC100 does the best on stance detection and QA, and EusCrawl does the best on NER. Wikipedia lags behind them all by a large margin. It is worth noting that the variance is high in certain tasks, which we attribute to the small size of the test sets and their unbalanced nature, but the general trends are consistent.

These results suggest that corpus quality issues in low-resource languages do not have a a major impact on NLU performance. Instead, we find evidence that it is the size and domain of the training corpus that is more important. This would explain why Wikipedia obtains the worst results, as it is substantially smaller than the other corpora and restricted to a narrow domain. Similarly, this is also consistent with EusCrawl performing worse than mC4 and CC100 on sentiment analysis and stance detection, as the domain of these benchmarks (tweets) is different from the domain of EusCrawl (primarily news, see Table \ref{tab:euscrawl_subsets}), while CommonCrawl-derived corpora are presumably more diverse.

Finally, we find that scaling to RoBERTa-large brings consistent improvements in all tasks. Thanks to this, we are able to outperform the best published results in all the 5 benchmarks. Note that we achieve this pre-training exclusively on Creative Commons data that we release publicly, while prior work relied on private datasets.

\section{Conclusions}

Taking Basque as a case study, our work gives further evidence that CommonCrawl-derived corpora have major quality issues in low-resource languages. At the same time, we show that ad-hoc crawling websites with high-quality content can be an effective alternative to collect data in such languages. Our resulting corpus EusCrawl has a higher quality than mC4 and CC100 according to our manual data audit, while being similar in size. Nevertheless, this improvement in quality does not carry over to downstream performance on NLU tasks, where we find evidence that data quantity and domain coverage are more important factors.

Our work leaves important lessons for future efforts on low-resource languages. First of all, we find that, even if CommonCrawl derived multilingual corpora do have major quality issues as raised by prior work \citep{kreutzer2021quality}, these issues do not have a significant impact in NLU tasks. This suggests that investing on bigger and more diverse datasets might be more fruitful than addressing such quality issues in low-resource settings. Given that the amount of written text in such languages is ultimately limited, we believe that developing effective cross-lingual transfer methods to exploit multilingual data is a promising future direction. Having said that, it should be noted that our study is limited to NLU tasks in a single language. It is possible that data quality plays a more important role in generation tasks, which we leave for future work to study. In addition, we think that it would be valuable to conduct similar studies in other languages to corroborate our findings.

Finally, we note that prior work on Basque NLP has often relied on private resources \citep{agerri-EtAl:2020:LREC}. Our work sets a new state-of-the-art on a diverse set of NLU benchmarks, and it does so using public data alone. By releasing our corpus, we hope to facilitate future work in Basque NLP, and encourage open and reproducible science using public resources.

\section*{Limitations}

Our evaluation focuses on NLU tasks, and it is possible that data quality plays a different role in generation tasks. We note, however, that generation quality is harder to evaluate through automatic metrics, which is why we decided to focus on NLU tasks. Moreover, the corpora that we compare differ on various aspects other than the data quality (e.g., the domain), and it is hard to isolate the effect of quality from the rest. In any case, we believe that our main claim still holds, in that data quality has a minor impact relative to such other factors. Finally, our work builds on EusCrawl---a new high-quality corpus that we introduce for Basque---and our analysis is thus limited to this language. It would be interesting to collect high-quality corpora for other low-resource languages, and conduct a similar comparison to corroborate that our findings also apply more broadly.

\section*{Acknowledgments}

We would like to thank Nikolas Vicuña for his help on expanding EusCrawl, Naman Goyal for his advice on pre-training RoBERTa, and Arantza Rico and Paloma Rodriguez-Miñambres for their help with the manual data audit. We are also grateful to all Basque media and content creators that share their work under a Creative Commons license, making a resource like EusCrawl possible.

Itziar Aldabe, Rodrigo Agerri, Olatz Perez-de-Viñaspre and Aitor Soroa were supported by the Basque Government (excellence research group IT1343-19 and DeepText project KK-2020/00088), and by the projects: (i) DeepKnowledge (PID2021-127777OB-C21) funded by MCIN/AEI/10.13039/501100011033 and FEDER Una manera de hacer Europa; (ii) Disargue (TED2021-130810B-C21), MCIN/AEI /10.13039/501100011033 and European Union NextGenerationEU/PRTR; (iii) DeepR3 (TED2021-130295B-C31) funded by MCIN/AEI /10.13039/501100011033 and EU NextGeneration programme EU/PRTR. Rodrigo Agerri's work is also supported by the RYC-2017-23647 fellowship (MCIN/AEI /10.13039/501100011033 y por El FSE invierte en tu futuro).

\bibliography{anthology,custom}

\begin{thebibliography}{17}
\expandafter\ifx\csname natexlab\endcsname\relax\def\natexlab#1{#1}\fi

\bibitem[{Agerri et~al.(2021)Agerri, Centeno, Espinosa, Fernandez~de Landa, and
  Rodrigo~Yuste}]{vaxxstance2021}
Rodrigo Agerri, Roberto Centeno, María Espinosa, Joseba Fernandez~de Landa,
  and Álvaro Rodrigo~Yuste. 2021.
\newblock Vaxxstance@iberlef 2021: Overview of the task on going beyond text in
  cross-lingual stance detection.

\bibitem[{Agerri et~al.(2020)Agerri, San~Vicente, Campos, Barrena, Saralegi,
  Soroa, and Agirre}]{agerri-EtAl:2020:LREC}
Rodrigo Agerri, Iñaki San~Vicente, Jon~Ander Campos, Ander Barrena, Xabier
  Saralegi, Aitor Soroa, and Eneko Agirre. 2020.
\newblock \href {https://www.aclweb.org/anthology/2020.lrec-1.588} {Give your
  text representation models some love: the case for basque}.
\newblock In \emph{Proceedings of The 12th Language Resources and Evaluation
  Conference}, pages 4781--4788, Marseille, France. European Language Resources
  Association.

\bibitem[{Alegria et~al.(2006)Alegria, Arregi, Ezeiza, and
  Fern{\'a}ndez}]{alegria2006lessons}
I{\~n}aki Alegria, Olatz Arregi, Nerea Ezeiza, and Izaskun Fern{\'a}ndez. 2006.
\newblock \href
  {http://journal.sepln.org/sepln/ojs/ojs/index.php/pln/article/view/2817}
  {Lessons from the development of a named entity recognizer for {B}asque}.
\newblock \emph{Procesamiento del Lenguaje Natural}, 36:25--37.

\bibitem[{Artetxe et~al.(2020)Artetxe, Ruder, and
  Yogatama}]{artetxe-etal-2020-cross}
Mikel Artetxe, Sebastian Ruder, and Dani Yogatama. 2020.
\newblock \href {https://doi.org/10.18653/v1/2020.acl-main.421} {On the
  cross-lingual transferability of monolingual representations}.
\newblock In \emph{Proceedings of the 58th Annual Meeting of the Association
  for Computational Linguistics}, pages 4623--4637, Online. Association for
  Computational Linguistics.

\bibitem[{Bommasani et~al.(2021)Bommasani, Hudson, Adeli, Altman, Arora, von
  Arx, Bernstein, Bohg, Bosselut, Brunskill, Brynjolfsson, Buch, Card,
  Castellon, Chatterji, Chen, Creel, Davis, Demszky, Donahue, Doumbouya,
  Durmus, Ermon, Etchemendy, Ethayarajh, Fei-Fei, Finn, Gale, Gillespie, Goel,
  Goodman, Grossman, Guha, Hashimoto, Henderson, Hewitt, Ho, Hong, Hsu, Huang,
  Icard, Jain, Jurafsky, Kalluri, Karamcheti, Keeling, Khani, Khattab, Koh,
  Krass, Krishna, Kuditipudi, Kumar, Ladhak, Lee, Lee, Leskovec, Levent, Li,
  Li, Ma, Malik, Manning, Mirchandani, Mitchell, Munyikwa, Nair, Narayan,
  Narayanan, Newman, Nie, Niebles, Nilforoshan, Nyarko, Ogut, Orr,
  Papadimitriou, Park, Piech, Portelance, Potts, Raghunathan, Reich, Ren, Rong,
  Roohani, Ruiz, Ryan, Ré, Sadigh, Sagawa, Santhanam, Shih, Srinivasan,
  Tamkin, Taori, Thomas, Tramèr, Wang, Wang, Wu, Wu, Wu, Xie, Yasunaga, You,
  Zaharia, Zhang, Zhang, Zhang, Zhang, Zheng, Zhou, and
  Liang}]{bommasani2021opportunities}
Rishi Bommasani, Drew~A. Hudson, Ehsan Adeli, Russ Altman, Simran Arora, Sydney
  von Arx, Michael~S. Bernstein, Jeannette Bohg, Antoine Bosselut, Emma
  Brunskill, Erik Brynjolfsson, Shyamal Buch, Dallas Card, Rodrigo Castellon,
  Niladri Chatterji, Annie Chen, Kathleen Creel, Jared~Quincy Davis, Dora
  Demszky, Chris Donahue, Moussa Doumbouya, Esin Durmus, Stefano Ermon, John
  Etchemendy, Kawin Ethayarajh, Li~Fei-Fei, Chelsea Finn, Trevor Gale, Lauren
  Gillespie, Karan Goel, Noah Goodman, Shelby Grossman, Neel Guha, Tatsunori
  Hashimoto, Peter Henderson, John Hewitt, Daniel~E. Ho, Jenny Hong, Kyle Hsu,
  Jing Huang, Thomas Icard, Saahil Jain, Dan Jurafsky, Pratyusha Kalluri,
  Siddharth Karamcheti, Geoff Keeling, Fereshte Khani, Omar Khattab, Pang~Wei
  Koh, Mark Krass, Ranjay Krishna, Rohith Kuditipudi, Ananya Kumar, Faisal
  Ladhak, Mina Lee, Tony Lee, Jure Leskovec, Isabelle Levent, Xiang~Lisa Li,
  Xuechen Li, Tengyu Ma, Ali Malik, Christopher~D. Manning, Suvir Mirchandani,
  Eric Mitchell, Zanele Munyikwa, Suraj Nair, Avanika Narayan, Deepak
  Narayanan, Ben Newman, Allen Nie, Juan~Carlos Niebles, Hamed Nilforoshan,
  Julian Nyarko, Giray Ogut, Laurel Orr, Isabel Papadimitriou, Joon~Sung Park,
  Chris Piech, Eva Portelance, Christopher Potts, Aditi Raghunathan, Rob Reich,
  Hongyu Ren, Frieda Rong, Yusuf Roohani, Camilo Ruiz, Jack Ryan, Christopher
  Ré, Dorsa Sadigh, Shiori Sagawa, Keshav Santhanam, Andy Shih, Krishnan
  Srinivasan, Alex Tamkin, Rohan Taori, Armin~W. Thomas, Florian Tramèr,
  Rose~E. Wang, William Wang, Bohan Wu, Jiajun Wu, Yuhuai Wu, Sang~Michael Xie,
  Michihiro Yasunaga, Jiaxuan You, Matei Zaharia, Michael Zhang, Tianyi Zhang,
  Xikun Zhang, Yuhui Zhang, Lucia Zheng, Kaitlyn Zhou, and Percy Liang. 2021.
\newblock \href {http://arxiv.org/abs/2108.07258} {On the opportunities and
  risks of foundation models}.

\bibitem[{Conneau et~al.(2020)Conneau, Khandelwal, Goyal, Chaudhary, Wenzek,
  Guzm{\'a}n, Grave, Ott, Zettlemoyer, and
  Stoyanov}]{conneau-etal-2020-unsupervised}
Alexis Conneau, Kartikay Khandelwal, Naman Goyal, Vishrav Chaudhary, Guillaume
  Wenzek, Francisco Guzm{\'a}n, Edouard Grave, Myle Ott, Luke Zettlemoyer, and
  Veselin Stoyanov. 2020.
\newblock \href {https://doi.org/10.18653/v1/2020.acl-main.747} {Unsupervised
  cross-lingual representation learning at scale}.
\newblock In \emph{Proceedings of the 58th Annual Meeting of the Association
  for Computational Linguistics}, pages 8440--8451, Online. Association for
  Computational Linguistics.

\bibitem[{Conneau and Lample(2019)}]{conneau2019xlm}
Alexis Conneau and Guillaume Lample. 2019.
\newblock \href
  {https://proceedings.neurips.cc/paper/2019/file/c04c19c2c2474dbf5f7ac4372c5b9af1-Paper.pdf}
  {Cross-lingual language model pretraining}.
\newblock In \emph{Advances in Neural Information Processing Systems},
  volume~32. Curran Associates, Inc.

\bibitem[{Kaplan et~al.(2020)Kaplan, McCandlish, Henighan, Brown, Chess, Child,
  Gray, Radford, Wu, and Amodei}]{kaplan2020scaling}
Jared Kaplan, Sam McCandlish, Tom Henighan, Tom~B. Brown, Benjamin Chess, Rewon
  Child, Scott Gray, Alec Radford, Jeffrey Wu, and Dario Amodei. 2020.
\newblock \href {http://arxiv.org/abs/2001.08361} {Scaling laws for neural
  language models}.

\bibitem[{Kreutzer et~al.(2021)Kreutzer, Caswell, Wang, Wahab, van Esch,
  Ulzii-Orshikh, Tapo, Subramani, Sokolov, Sikasote, Setyawan, Sarin, Samb,
  Sagot, Rivera, Rios, Papadimitriou, Osei, Suárez, Orife, Ogueji, Rubungo,
  Nguyen, Müller, Müller, Muhammad, Muhammad, Mnyakeni, Mirzakhalov,
  Matangira, Leong, Lawson, Kudugunta, Jernite, Jenny, Firat, Dossou, Dlamini,
  de~Silva, Çabuk Ballı, Biderman, Battisti, Baruwa, Bapna, Baljekar, Azime,
  Awokoya, Ataman, Ahia, Ahia, Agrawal, and Adeyemi}]{kreutzer2021quality}
Julia Kreutzer, Isaac Caswell, Lisa Wang, Ahsan Wahab, Daan van Esch,
  Nasanbayar Ulzii-Orshikh, Allahsera Tapo, Nishant Subramani, Artem Sokolov,
  Claytone Sikasote, Monang Setyawan, Supheakmungkol Sarin, Sokhar Samb,
  Benoît Sagot, Clara Rivera, Annette Rios, Isabel Papadimitriou, Salomey
  Osei, Pedro~Ortiz Suárez, Iroro Orife, Kelechi Ogueji, Andre~Niyongabo
  Rubungo, Toan~Q. Nguyen, Mathias Müller, André Müller, Shamsuddeen~Hassan
  Muhammad, Nanda Muhammad, Ayanda Mnyakeni, Jamshidbek Mirzakhalov,
  Tapiwanashe Matangira, Colin Leong, Nze Lawson, Sneha Kudugunta, Yacine
  Jernite, Mathias Jenny, Orhan Firat, Bonaventure F.~P. Dossou, Sakhile
  Dlamini, Nisansa de~Silva, Sakine Çabuk Ballı, Stella Biderman, Alessia
  Battisti, Ahmed Baruwa, Ankur Bapna, Pallavi Baljekar, Israel~Abebe Azime,
  Ayodele Awokoya, Duygu Ataman, Orevaoghene Ahia, Oghenefego Ahia, Sweta
  Agrawal, and Mofetoluwa Adeyemi. 2021.
\newblock \href {http://arxiv.org/abs/2103.12028} {Quality at a glance: An
  audit of web-crawled multilingual datasets}.

\bibitem[{Kudo and Richardson(2018)}]{kudo-richardson-2018-sentencepiece}
Taku Kudo and John Richardson. 2018.
\newblock \href {https://doi.org/10.18653/v1/D18-2012} {{S}entence{P}iece: A
  simple and language independent subword tokenizer and detokenizer for neural
  text processing}.
\newblock In \emph{Proceedings of the 2018 Conference on Empirical Methods in
  Natural Language Processing: System Demonstrations}, pages 66--71, Brussels,
  Belgium. Association for Computational Linguistics.

\bibitem[{Lai et~al.(2021)Lai, Cignarella, Finos, and
  Sciandra}]{Lai2021WordUpAV}
Mirko Lai, Alessandra~Teresa Cignarella, Livio Finos, and Andrea Sciandra.
  2021.
\newblock Wordup! at vaxxstance 2021: Combining contextual information with
  textual and dependency-based syntactic features for stance detection.
\newblock In \emph{IberLEF@SEPLN}.

\bibitem[{Lin et~al.(2021)Lin, Mihaylov, Artetxe, Wang, Chen, Simig, Ott,
  Goyal, Bhosale, Du, Pasunuru, Shleifer, Koura, Chaudhary, O'Horo, Wang,
  Zettlemoyer, Kozareva, Diab, Stoyanov, and Li}]{lin2021fewshot}
Xi~Victoria Lin, Todor Mihaylov, Mikel Artetxe, Tianlu Wang, Shuohui Chen,
  Daniel Simig, Myle Ott, Naman Goyal, Shruti Bhosale, Jingfei Du, Ramakanth
  Pasunuru, Sam Shleifer, Punit~Singh Koura, Vishrav Chaudhary, Brian O'Horo,
  Jeff Wang, Luke Zettlemoyer, Zornitsa Kozareva, Mona Diab, Veselin Stoyanov,
  and Xian Li. 2021.
\newblock \href {http://arxiv.org/abs/2112.10668} {Few-shot learning with
  multilingual language models}.

\bibitem[{Liu et~al.(2019)Liu, Ott, Goyal, Du, Joshi, Chen, Levy, Lewis,
  Zettlemoyer, and Stoyanov}]{liu2019roberta}
Yinhan Liu, Myle Ott, Naman Goyal, Jingfei Du, Mandar Joshi, Danqi Chen, Omer
  Levy, Mike Lewis, Luke Zettlemoyer, and Veselin Stoyanov. 2019.
\newblock \href {http://arxiv.org/abs/1907.11692} {Roberta: A robustly
  optimized bert pretraining approach}.

\bibitem[{Otegi et~al.(2020)Otegi, Agirre, Campos, Soroa, and
  Agirre}]{otegi-EtAl:2020:LREC}
Arantxa Otegi, Aitor Agirre, Jon~Ander Campos, Aitor Soroa, and Eneko Agirre.
  2020.
\newblock \href {https://www.aclweb.org/anthology/2020.lrec-1.55}
  {Conversational question answering in low resource scenarios: A dataset and
  case study for basque}.
\newblock In \emph{Proceedings of The 12th Language Resources and Evaluation
  Conference}, pages 436--442, Marseille, France. European Language Resources
  Association.

\bibitem[{Pires et~al.(2019)Pires, Schlinger, and
  Garrette}]{pires-etal-2019-multilingual}
Telmo Pires, Eva Schlinger, and Dan Garrette. 2019.
\newblock \href {https://doi.org/10.18653/v1/P19-1493} {How multilingual is
  multilingual {BERT}?}
\newblock In \emph{Proceedings of the 57th Annual Meeting of the Association
  for Computational Linguistics}, pages 4996--5001, Florence, Italy.
  Association for Computational Linguistics.

\bibitem[{Rae et~al.(2022)Rae, Borgeaud, Cai, Millican, Hoffmann, Song,
  Aslanides, Henderson, Ring, Young, Rutherford, Hennigan, Menick, Cassirer,
  Powell, van~den Driessche, Hendricks, Rauh, Huang, Glaese, Welbl, Dathathri,
  Huang, Uesato, Mellor, Higgins, Creswell, McAleese, Wu, Elsen, Jayakumar,
  Buchatskaya, Budden, Sutherland, Simonyan, Paganini, Sifre, Martens, Li,
  Kuncoro, Nematzadeh, Gribovskaya, Donato, Lazaridou, Mensch, Lespiau,
  Tsimpoukelli, Grigorev, Fritz, Sottiaux, Pajarskas, Pohlen, Gong, Toyama,
  de~Masson~d'Autume, Li, Terzi, Mikulik, Babuschkin, Clark, de~Las~Casas, Guy,
  Jones, Bradbury, Johnson, Hechtman, Weidinger, Gabriel, Isaac, Lockhart,
  Osindero, Rimell, Dyer, Vinyals, Ayoub, Stanway, Bennett, Hassabis,
  Kavukcuoglu, and Irving}]{rae2022scaling}
Jack~W. Rae, Sebastian Borgeaud, Trevor Cai, Katie Millican, Jordan Hoffmann,
  Francis Song, John Aslanides, Sarah Henderson, Roman Ring, Susannah Young,
  Eliza Rutherford, Tom Hennigan, Jacob Menick, Albin Cassirer, Richard Powell,
  George van~den Driessche, Lisa~Anne Hendricks, Maribeth Rauh, Po-Sen Huang,
  Amelia Glaese, Johannes Welbl, Sumanth Dathathri, Saffron Huang, Jonathan
  Uesato, John Mellor, Irina Higgins, Antonia Creswell, Nat McAleese, Amy Wu,
  Erich Elsen, Siddhant Jayakumar, Elena Buchatskaya, David Budden, Esme
  Sutherland, Karen Simonyan, Michela Paganini, Laurent Sifre, Lena Martens,
  Xiang~Lorraine Li, Adhiguna Kuncoro, Aida Nematzadeh, Elena Gribovskaya,
  Domenic Donato, Angeliki Lazaridou, Arthur Mensch, Jean-Baptiste Lespiau,
  Maria Tsimpoukelli, Nikolai Grigorev, Doug Fritz, Thibault Sottiaux, Mantas
  Pajarskas, Toby Pohlen, Zhitao Gong, Daniel Toyama, Cyprien
  de~Masson~d'Autume, Yujia Li, Tayfun Terzi, Vladimir Mikulik, Igor
  Babuschkin, Aidan Clark, Diego de~Las~Casas, Aurelia Guy, Chris Jones, James
  Bradbury, Matthew Johnson, Blake Hechtman, Laura Weidinger, Iason Gabriel,
  William Isaac, Ed~Lockhart, Simon Osindero, Laura Rimell, Chris Dyer, Oriol
  Vinyals, Kareem Ayoub, Jeff Stanway, Lorrayne Bennett, Demis Hassabis, Koray
  Kavukcuoglu, and Geoffrey Irving. 2022.
\newblock \href {http://arxiv.org/abs/2112.11446} {Scaling language models:
  Methods, analysis \& insights from training gopher}.

\bibitem[{Xue et~al.(2021)Xue, Constant, Roberts, Kale, Al-Rfou, Siddhant,
  Barua, and Raffel}]{xue-etal-2021-mt5}
Linting Xue, Noah Constant, Adam Roberts, Mihir Kale, Rami Al-Rfou, Aditya
  Siddhant, Aditya Barua, and Colin Raffel. 2021.
\newblock \href {https://doi.org/10.18653/v1/2021.naacl-main.41} {m{T}5: A
  massively multilingual pre-trained text-to-text transformer}.
\newblock In \emph{Proceedings of the 2021 Conference of the North American
  Chapter of the Association for Computational Linguistics: Human Language
  Technologies}, pages 483--498, Online. Association for Computational
  Linguistics.

\end{thebibliography}
\bibliographystyle{acl_natbib}

\appendix

\section{Annotation instructions} \label{app:instructions}

\begin{table*}[ht]
\begin{center}
\begin{small}
\begin{tabular}{cp{13cm}}
\toprule
\multirow{5}{*}{\shortstack{LangID}}
& EGOKIA: Dokumentua euskaraz dago. \\
& \textit{CORRECT: The document is in Basque.} \\
\cmidrule{2-2}
& ARAZOAK: Dokumentuaren zati esanguratsu bat ez dago euskaraz. \\
& \textit{PROBLEMATIC: A significant portion of the document is not in Basque.} \\

\midrule
\multirow{6}{*}{\shortstack{Hizkuntza \\  \\ \textit{Lang. variety}}}
& EGOKIA: Dokumentua hizkuntza estandar eta zuzenean idatzia dago. \\
& \textit{CORRECT: The document is written in standard and correct language.} \\
\cmidrule{2-2}
& ARAZOAK: Dokumentua ez dago hizkuntza estandar edo zuzenean idatzia (adb. euskalkiren batean dago ala itzulpen automatikoaren bidez sortua dirudi). \\
& \textit{PROBLEMATIC: The document is not written in standard and correct language (e.g., it is written in a dialect using non-standard Basque, or it seems to be generated through machine translation).} \\

\midrule
\multirow{6}{*}{\shortstack{Koherentzia \\  \\ \textit{Coherence}}}
& EGOKIA: Dokumentua koherentea da, eta hasieratik bukaerara unitate bat osatzen du. \\
& \textit{CORRECT: The document is coherent, and it constitutes a single unit from the beginning to the end.} \\
\cmidrule{2-2}
& ARAZOAK: Dokumentua ez da koherentea: hutsuneak ditu edota atal batzuk ez dute elkarren artean loturarik (dokumentu ezberdinak dirudite). \\
& \textit{PROBLEMATIC: The document is not coherent: it has gaps and/or some portions do not seem connected (they seem to come from separate documents).} \\

\midrule
\multirow{6}{*}{\shortstack{Garbitasuna \\  \\ \textit{Noise}}}
& EGOKIA: Dokumentuko testua garbia da. \\
& \textit{CORRECT: The text in the document is clean.} \\
\cmidrule{2-2}
& ARAZOAK: Dokumentua ez da erabat garbia, eta benetako testuaz gain webguneko bestelako elementuak daude (menuetako testua, html kodea...). \\
& \textit{PROBLEMATIC: The document is not entirely clean, and there are other elements in addition to the real content (text from menus, HTML code...).} \\

\midrule
\multirow{6}{*}{\shortstack{Edukia \\  \\ \textit{Content}}}
& EGOKIA: Dokumentua pertsona batek sortua dirudi eta gutxieneko mami bat du. \\
& \textit{CORRECT: The document seems to have been created by a human and has some minimum meat.} \\
\cmidrule{2-2}
& ARAZOAK: Dokumentuak automatikoki sortua dirudi edota ez du inolako mamirik (adb futbol ligako sailkapen-taula). \\
& \textit{The document seems to have been generated automatically and/or has no meat at all (e.g., a soccer standing table).} \\

\midrule
\multirow{8}{*}{\shortstack{Kalitate orokorra \\  \\ \textit{Overall quality}}}
& ALTUA: Dokumentua kalitatezkoa da, eta corpusean izatea komeniko litzatekeela uste dut. \\
& \textit{HIGH: The document is of good quality, and I think that it would be good to have it in the corpus.} \\
\cmidrule{2-2}
& ERTAINA: Dokumentuak arazo batzuk ditu baina ez dira larriak, eta ez nago ziur ea corpusean izatea komeniko litzatekeen. \\
& \textit{MEDIUM: The document has minor issues, and I am not sure if it would be good to have it in the corpus.} \\
\cmidrule{2-2}
& BAXUA: Dokumentuak arazo nabarmenak ditu. Ez dut uste corpusean izatea komeniko litzatekeenik. \\
& \textit{LOW: The document has major issues. I think that it would be better not to have it in the corpus.} \\

\bottomrule
\end{tabular}%
\end{small}
\end{center}
\caption{Annotation instructions used for the qualitative evaluation. We report the original instructions in Basque, as well as the corresponding translation into English.}
\label{tab:instructions}
\end{table*}

Table \ref{tab:instructions} reports the complete instructions used for the qualitative evaluation as given to the annotators.

\section{Downstream evaluation} \label{app:downstream_settings}

We next provide additional details on the datasets used for downstream evaluation:
\begin{itemize}
\item \textbf{Topic classification}: The Basque Headlines Topic Classification (BHTC) dataset \cite{agerri-EtAl:2020:LREC} contains 12k headlines from the Argia news magazine classified into 12 thematic categories\footnote{\scriptsize{\url{https://hizkuntzateknologiak.elhuyar.eus/assets/files/bhtc.tgz}}}. We use the standard splits containing 8662 examples for training, 1861 for development and 1860 for testing.

\item \textbf{Sentiment classification}: The Behagune dataset\footnote{\scriptsize{\url{https://hizkuntzateknologiak.elhuyar.eus/assets/files/behaguneadss2016-dataset.tgz}}} comprises 2936 tweets in Basque labeled as positive, negative or neutral. We used the same splits for train (80\%), test (10\%) and development (10\%) as in \citet{agerri-EtAl:2020:LREC}.
\item \textbf{Stance detection}: We used the VaxxStance dataset \cite{vaxxstance2021}, which offers tweets labeled as expressing an AGAINST, FAVOR or NEUTRAL stance with respect to vaccines. It contains 1070 tweets for training and 313 for testing\footnote{\scriptsize{\url{https://vaxxstance.github.io/}}}. 
\item \textbf{Named Entity Recognition (NER)}: EIEC\footnote{\scriptsize{\url{http://ixa2.si.ehu.eus/eiec/eiec_v1.0.tgz}}} \citep{alegria2006lessons} is a Basque NER dataset composed of 44K training tokens ($3817$ unique entities) and 15K test tokens ($931$ entities).
\item \textbf{Question Answering (QA)}: Elkarrizketak is an extractive conversational QA dataset \cite{otegi-EtAl:2020:LREC} that contains 377 dialogues (301 train, 38 development and 38 test) and 1,634 question/answer pairs (1,306 train, 161 development and 167 test)\footnote{\scriptsize{\url{http://ixa.si.ehu.es/node/12934}}}. %
\end{itemize}

\end{document}